\documentclass[letterpaper, 10 pt, conference]{ieeeconf}  

\IEEEoverridecommandlockouts                              

\usepackage{cite}
\usepackage{amsmath,amssymb,amsfonts}
\usepackage{algorithmic}
\usepackage{graphicx}
\usepackage{textcomp}
\usepackage{subfigure}
\usepackage{booktabs} 
\usepackage{siunitx} 

\usepackage{array} 
\usepackage[table]{xcolor} 
\usepackage{multirow} 

\title{\LARGE \bf
Learning secondary tool affordances of human partners using iCub robot's egocentric data
}

\author{
Bosong Ding$^{1}$, Erhan Oztop$^{2}$, Giacomo Spigler$^{1}$, Murat Kirtay$^{1}$
\thanks{$^{1}$Bosong Ding is with the department of Cognitive Science and Artificial Intelligence, Tilburg University, Tilburg, Netherlands.
        {\tt\small \{b.ding\_3\}@tilburguniversity.edu}}%
\thanks{$^{2}$Erhan Oztop is with Symbiotic Intelligent Systems Research Center, Institute for Open and Transdisciplinary, Research Initiatives, Osaka University, Japan. He is also affiliated with  Ozyegin University, Istanbul, Turkey.
        {\tt\small \{erhan.oztop\}@otri.osaka-u.ac.jp}}%
\thanks{$^{1}$Giacomo Spigler is with the department of Cognitive Science and Artificial Intelligence, Tilburg University, Tilburg, Netherlands.
        {\tt\small \{g.spigler\}@tilburguniversity.edu}}%
\thanks{$^{1}$Murat Kirtay is with the department of Cognitive Science and Artificial Intelligence, Tilburg University, Tilburg, Netherlands.
        {\tt\small \{m.kirtay\}@tilburguniversity.edu}}%
}

\begin{document}

\maketitle
\thispagestyle{empty}
\pagestyle{empty}

\begin{abstract}

Objects, in particular tools, provide several action possibilities to the agents that can act on them, which are generally associated with the term of affordances. A tool is typically designed for a specific purpose, such as driving a nail in the case of a hammer, which we call as the primary affordance. A tool can also be used beyond its primary purpose, in  which case we can associate this auxiliary use with the term secondary affordance. Previous work on affordance perception and learning has been mostly focused on primary affordances.  Here, we address the less explored problem of learning the secondary tool affordances of human partners. To do this, we use the iCub robot to observe human partners with three cameras while they perform actions on twenty objects using four different tools. In our experiments, human partners utilize tools to perform actions that do not correspond to their primary affordances. For example, the iCub robot observes a human partner using a ruler for pushing, pulling, and moving objects instead of measuring their lengths. In this setting, we constructed a dataset by taking images of objects before and after each action is executed. We then model learning secondary affordances by training three neural networks (ResNet-18, ResNet-50, and ResNet-101) each on three tasks, using raw images showing the `initial' and `final' position of objects as input: (1) predicting the tool used to move an object, (2) predicting the tool used with an additional categorical input that encoded the action performed, and (3) joint prediction of both tool used and action performed. Our results indicate that deep learning architectures enable the iCub robot to predict secondary tool affordances, thereby paving the road for human-robot collaborative object manipulation involving complex affordances.
\end{abstract}

\section{Introduction}
Humans and animals rely on perceiving the possible actions (affordances) a tool can offer to accomplish a wide array of tasks. These can be as simple as foraging for basic survival or as complex as creating a design for an industrial application that includes human-robot collaboration \cite{ye2009perceiving}, \cite{jamone2016affordances}, \cite{stramandinoli2019affordance}. While tools are generally crafted with a primary function in mind, they frequently possess the potential for other uses beyond their original intent. Take a ruler as an instance; it is primarily used to measure the size of an object, yet it can also serve to pull an object closer. The first use is an example of the tool's \emph{primary affordance}, while the second represents one of its \emph{secondary affordances} \cite{ye2009perceiving}. 

The concept of affordance has been explored across multiple fields, such as user interface design, human-computer interaction, and robotics. However, much of the research has focused on recognizing and utilizing primary affordances, especially in the field of robotics \cite{jamone2016affordances}, \cite{norman2013design}, \cite{masoudi2019review}.

In this study, we addressed one of the less explored areas of affordances in robotics, focusing on the ability of the iCub robot to learn about secondary tool functions by observing its human partners through an egocentric perspective. Specifically, we asked the following research question: \emph{How can a multisensory-enabled iCub robot learn the secondary affordances of tools used by human partners?} To explore this, we designed an interactive experiment to recognize secondary affordances of tools using image data acquired on the iCub robot via three cameras before and after operators act on various objects. It is important to note that in our study, tools are used by human partners in ways that go beyond their intended function. For example, \emph{throwability} is a primary affordance of the boomerang tool, but in our experiments, a boomerang was repurposed to perform various actions like pushing and pulling objects.

We assess the ability of Convolutional Neural Networks architectures (ResNet-18, ResNet-50, and ResNet-101) to predict the secondary affordances of tools across three tasks. The initial task involves identifying tools from raw images of the object captured before and after performing an action. The second task enhances this by including information about the action being executed. Finally, the third task aims to independently predict the tool and action based solely on the raw images.

Our results show that the models achieve significant accuracy ($>$ 90\%) while predicting tools and tool-action pairs across three tasks. Based on these findings, we put forward that the ResNet-based architectures, particularly ResNet-50, suit tool affordance learning using egocentric data. 

Our work offers the following contributions. Firstly, we conduct one of the first studies using the iCub robot to understand secondary affordances in human interactions within a real-world environment. Secondly, we deliver extensive benchmarking results and demonstrate the effectiveness of neural networks for learning secondary tool affordances.

The structure of this paper is as follows: Section~\ref{related_work} reviews the literature on affordance learning in robotics. Section~\ref{dataset} details the dataset specifics used in our research. The tasks we investigate are described in Section~\ref{tasks}. Our methods for developing a secondary affordance learning framework are explained in Section~\ref{methods}. Section~\ref{results} presents results, and Section~\ref{reprod} provides access to the online repository for replicating these results. Lastly, Section~\ref{conclusion} summarizes our conclusions and outlines potential directions for future work.

\section{Related Work} \label{related_work}

Tool usage has been one of the active research areas in robotics \cite{tee2018towards}, \cite{brown2011tool}. Due to its broad scope, here, we review the studies on tool affordance prediction and learning studies in robotics. We also note that most studies in robotics concern primary rather than secondary affordances \cite{jamone2016affordances}, \cite{zech2017computational}, \cite{min2016affordance}.

Varadarajan and Vince \cite{varadarajan2012afrob} present the extension of affordance computing initiative (AfNet) to robotics applications, AfRob, to employ vision data to recognize object affordances, such as containability, rollability, grab supportability, etc.  Although the motivation of the study was framed for social robots in a domestic environment, the authors did not provide an application for a real robot.  Chu et al. \cite{chu2019toward} employ a manipulator with RGB-D sensor to detect and rank objects' multi-affordances: primary (e.g., grasping a hammer from handle) and non-primary (e.g., grasping a hammer from head) affordances of objects. The authors designed a convolutional neural network  (CNN) architecture with an RGB-D data processing pipeline to detect affordances by labeling each image pixel for different affordance types.  Lai et al. \cite{lai2021functional} used a deep learning architecture to determine the regions of the different objects that are functionally the same (i.e., functional correspondence problem) in performing a task, such as the body of a bottle and the front part of a shoe can be used a task of pounding. Here, the authors draw parallels between the secondary affordances of the objects and functional correspondence. To realize their approach, the authors presented a dataset of RGB images for different objects with the same affordances (i.e., actions).  The CNN architecture was adopted to detect functionally corresponding points in images. Do et al. \cite{do2018affordancenet} proposed a convolutional network-based primary affordance and object detection framework.  The results show that the implementation outperforms some state-of-the-art models on affordance detection tasks using RGB images from off-the-shelf datasets. The authors provide a robotic application to offer the same approach that can be employed on a real robot platform to perform a task that requires grasping affordances, e.g., pouring a bottle. 
 
Based on the studies introduced above, we can conclude that our work offers the following differences. On the one hand, unlike existing studies that enable robots to detect object affordances, our study employs the iCub robot to learn human secondary affordances using egocentric data.  On the other hand, we achieve the results in a noisy real-world setting where the iCub robot is physically co-located with human partners.

\section{Dataset Acquisition Setup and Specifications} \label{dataset}

In this section, we describe the dataset acquisition procedure for learning the secondary affordances of human partners by employing the iCub humanoid robot~\cite{kirtay2020icub}.  

\begin{figure}[ht!]
	\centering
	\begin{center}
		\subfigure[Experiment setup]{\label{toolsAndObjs:a}\includegraphics[width=0.46\textwidth]{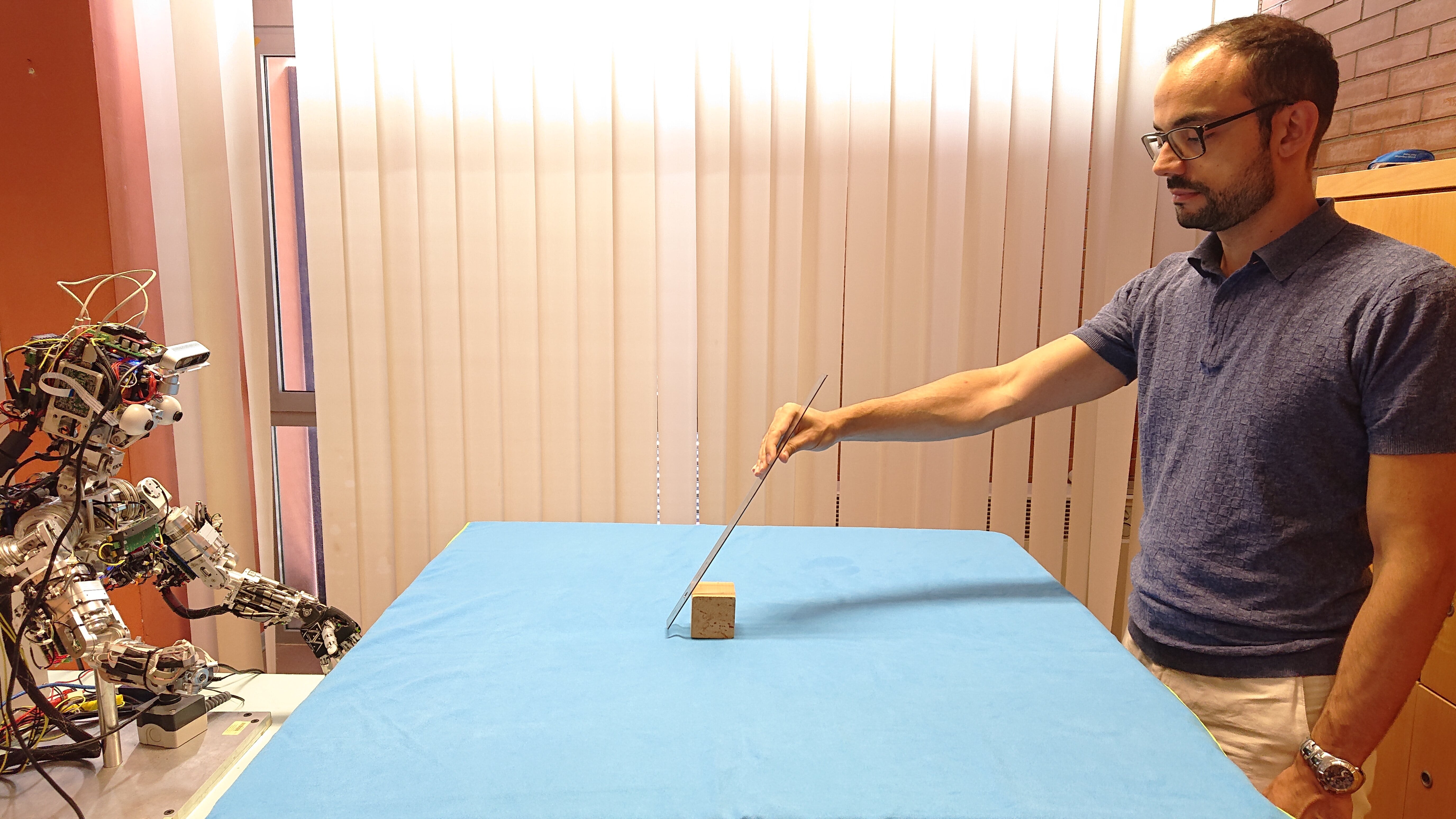}}\\
		\subfigure[Objects ]{\label{toolsAndObjs:b}\includegraphics[width=0.23\textwidth, height=0.15\textheight]{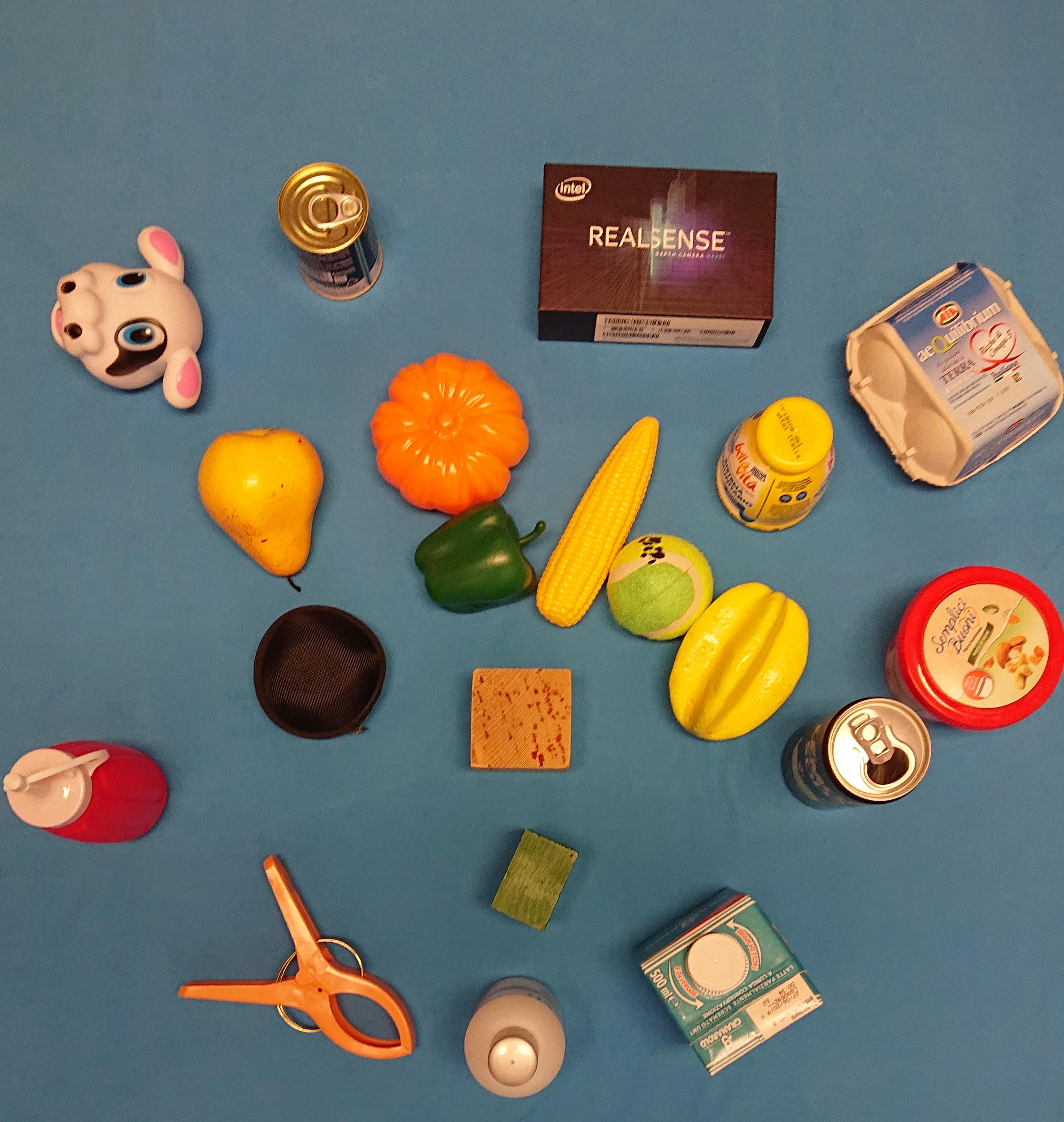}} 
		\subfigure[Tools ]{\label{toolsAndObjs:c}\includegraphics[width=0.23\textwidth, height=0.15\textheight]{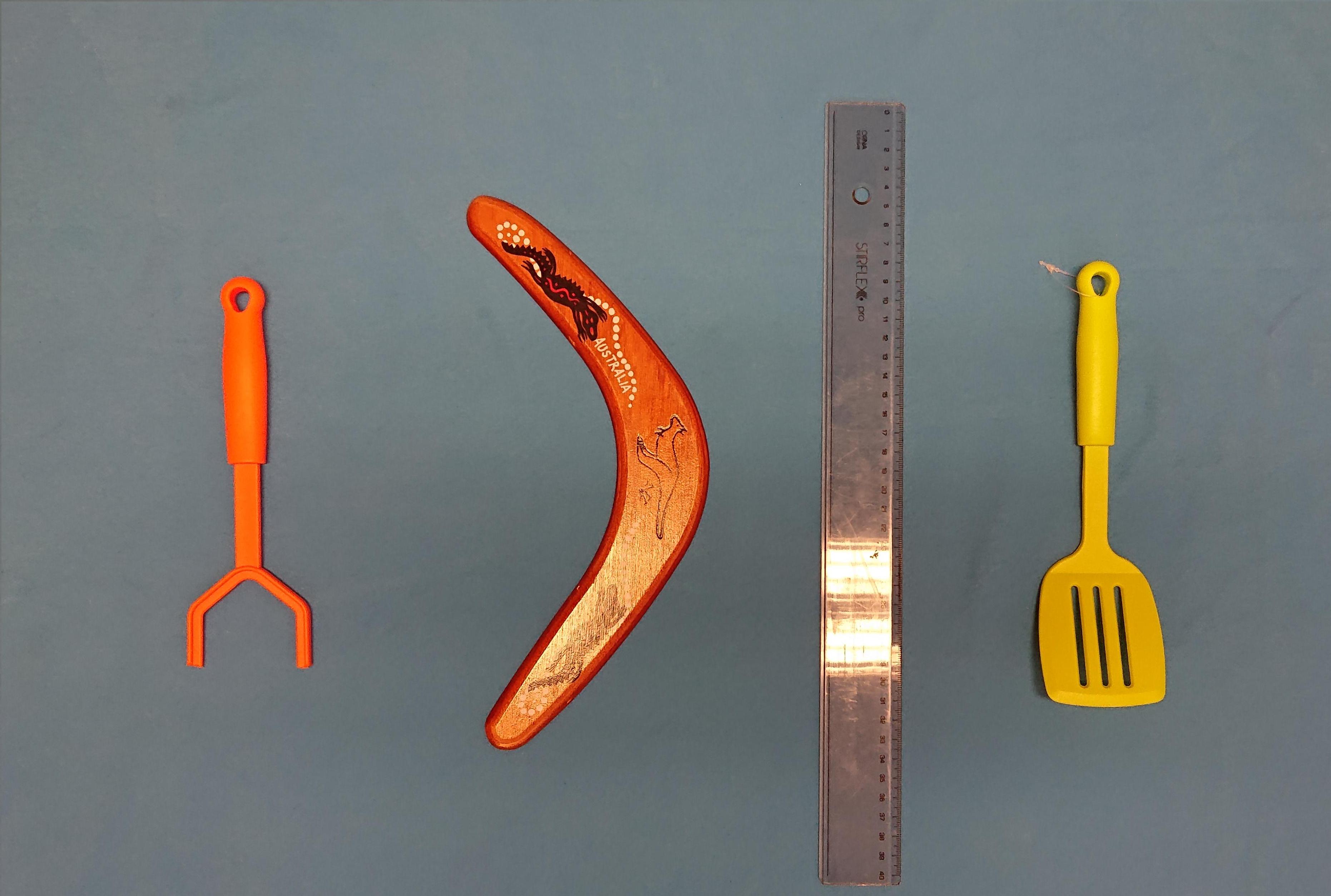}}
	\end{center}
	\caption{ The experimental setup (a), where an operator performs a pull action on a wooden cube with a ruler as a tool, the objects (b) and tools (c) that were employed to construct the dataset.}
	\label{fig:toolsAndObjs}
\end{figure}

Our experimental setup consists of the upper body of the iCub robot that has three color cameras (two Dragonfly cameras in each eye and an Intel Realsense d435i camera located above the eyes). Figure~\ref{toolsAndObjs:a} shows a photo from one of the experiments where a right-handed operator uses a ruler as a tool to perform an action that is associated with its secondary affordance, i.e., pulling a wooden cube with a ruler.
In this setting, we designed the experiment where the iCub robot becomes an action observer, and four different human partners (i.e., operators) are assigned to be action performers. To construct the dataset, we used  20 objects varying in material, color, size, shape, etc.

As shown in Figure~\ref{toolsAndObjs:a}, the operator performs predefined actions (namely, push, pull, left to right, and right to left) on the object using a tool. The objects and tools (boomerang, ruler, slingshot, and spatula) used in this experimental setup are shown in Figure~\ref{toolsAndObjs:b},~\ref{toolsAndObjs:c}, respectively. We note that the operators do not perform primary affordance of the tools. For example, see  Figure~\ref{toolsAndObjs:a}, the operator did not measure the size of an object with the ruler as a tool; instead, the ruler was used for performing predefined actions: pulling, pushing, moving an object from left to right, and moving an object from right to left.

\begin{figure}[ht!]
	\centering
	\begin{center}
		\subfigure[Initial pose of the object]{\label{toolsAndObjs:b_2nd}\includegraphics[width=0.235\textwidth, height=0.15\textheight]{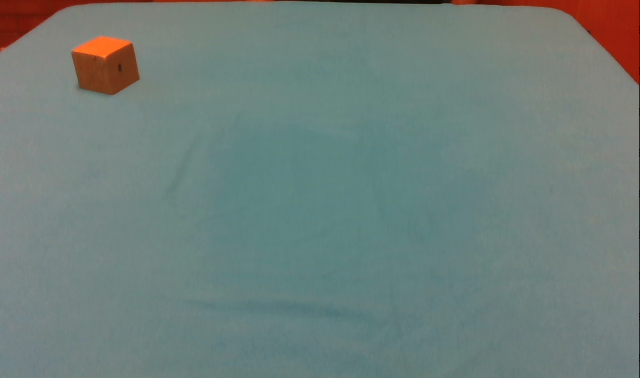}} 
		\subfigure[Final pose of the object ]{\label{toolsAndObjs:c_2nd}\includegraphics[width=0.235\textwidth, height=0.15\textheight]{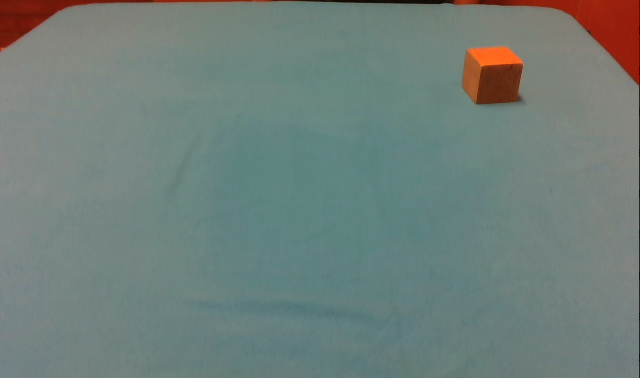}}
	\end{center}
	\caption{(a) Initial pose of the object and (b) final pose of the object after performing an action (left to right) with ruler as a tool.}
	\label{fig:init_effect}
\end{figure}

We captured images using three cameras to record the initial and final poses of objects before and after actions were performed. Figure \ref{fig:init_effect} illustrates the initial and final pose of a wooden cube object from one of the cameras located on the robot. Each object was subject to an action using a specific tool, and this process was repeated 10 times for every combination of object, action, and tool. After the experiment, we acquired a total of 3200 samples, each consisting of 6 color images with a resolution of $640\times480$ pixels (i.e., each sample one of all combinations of 20 objects, 4 tools, 4 actions, 10 repetitions, and consist of initial and final pairs of images for each of the three cameras).
 
We note that the robot's role in this setting is to observe human partners' actions using a set of tools. However, the results presented in this work can be used as a foundation for employing the iCub robot in a human-robot collaboration task where the robot actively interacts with the human partner and performs secondary affordances on the tools.

\section{Tasks} \label{tasks}

This paper focuses on two tasks: 1) tool recognition with and without actions as input and 2) predicting actions and tools that match the change in state of an object after manipulation by a human partner.


\subsection{Tool Recognition} 
In tool recognition task, the model is provided with initial and final images and information regarding the action used for object manipulation. The objective is for the model to predict the tools used for the observed action. Additionally, we conducted experiments using models that received explicit information about the action performed.

\subsection{Action and Tool Recognition} \label{toool_task}
This task involves feeding the models with initial and final images. The objective is to recognize the specific combination of tool and action responsible for transitioning between the initial and final images, determining the exact action and tools used in manipulating the object. With four tools and four actions, the goal is to identify the appropriate combination of action and tool accurately.

\section{Methods} \label{methods}
In this section, we explain the pre-processing of the dataset to create suitable input-output pairs for the tasks. Next, we discuss the network architectures employed for the tool affordance learning. We note that materials are made available for reproducibility of the study on our GitHub repository, see Section \ref{reprod}.

\begin{figure*}[ht!]
\centering
\includegraphics[width=0.92\textwidth]{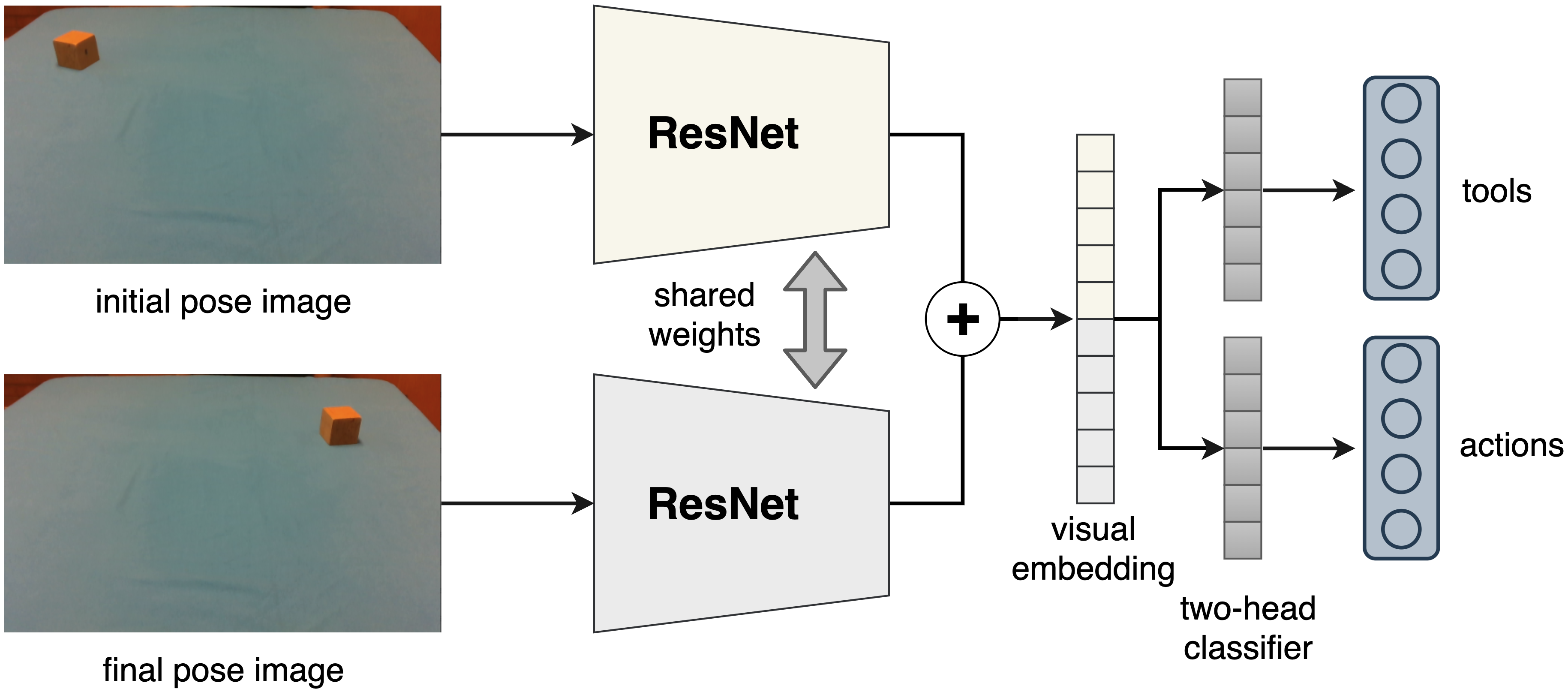}
\caption{Schematic representation of the ResNet-50 architecture for tool-action pair prediction described in Section \ref{toool_task}. The architecture uses paired initial and final images processed through shared-weight ResNet layers. The feature maps generated using two independent branches are merged and fed into a two-headed classifier to simultaneously predict the tools and actions.}
\label{fig:architecture}
\end{figure*}

\subsection{Data Pre-processing}
We explored the tasks introduced in Section \ref{tasks} using four different tools (boomerang, ruler, slingshot, and spatula) and four pre-defined actions (push, pull, left-to-right, right-to-left) across a range of 20 objects. For the tool prediction tasks, the goal is to predict one out of four possible tools. The action+tool recognition task is more complex, requiring the correct prediction of action and tool simultaneously. Each data sample includes `initial' and `final' RGB images, representing the scene before and after manipulation, respectively, taken from three robot-mounted egocentric cameras (center, left, right). These images are resized to $128\times128$ pixels and normalized. In the second task, we incorporate an extra input as a one-hot vector encoding the action taken. The dataset is split into training, validation, and test sets with a ratio of 6:2:2, across action-tool repetitions for each object.

\subsection{Action and Tool Recognition Models}
Action and tool recognition is performed by training Convolutional Neural Networks (CNNs) \cite{lecun2015deep} on the three target tasks due to their proven effectiveness on image classification problems. To benchmark the results, we compare three variants of the ResNet architecture \cite{he2016deep}. ResNet-18, comprising 11.7 million parameters; ResNet-50, with 25.6 million parameters; and ResNet-101, with 44.5 million parameters.  We implemented these ResNet architectures because the studies show that they are a particularly effective neural network architecture whose main feature is `skip-connection' that improves the flow of gradients across many layers without incurring the problem of vanishing gradients \cite{visualizing_loss_landscape}.

\subsection{Network Architectures}
We evaluate various ResNet-based network architectures on action and tool recognition tasks to select the most effective approach. We employed these networks to determine the differences between the initial and final configurations in each task. Therefore, we anticipate a significant variance in performance among models based on how they process these representations. Specifically, we investigate five distinct methods of handling the data samples. Here, \emph{C} and \emph{N} stand for camera and network, respectively.

\begin{enumerate}
    \item \emph{Stacked-channels (3C-1N)}: We stack the initial and final images from all three cameras (i.e., 2x3x3=18 image channels, as \{initial,final\}x\{left,center,right\}x\{red,green,blue\}), processed by a single ResNet. Although this is the simplest model we examine in this study, it might face issues due to the lack of pixel overlap among images from different cameras and varying initial and final states.

    \item \emph{Separate (3C-6N)}: Here, we treat each of the six images (initial and final from left, center, and right cameras) independently. Each image is processed by its own ResNet, leading to six distinct networks. The resulting embeddings are concatenated to form the input to a final common output layer, as illustrated in Fig. \ref{fig:architecture}.

    \item \emph{Separate-central (1C-2N)}: Only the central camera's initial and final images are considered, each being processed by a separate ResNet. This gives rise to two individual neural networks.

    \item \emph{Shared (3C-3N)}: In this configuration, each initial and final image is individually processed through the same network dedicated to each camera, using shared weights. This means we employ three networks, with each used twice. The embeddings from all networks are combined as in the previous models (see Fig. \ref{fig:architecture}).

    \item \emph{Shared-central (1C-1N)}: A single ResNet is used for both the initial and final images from just the central camera, leveraging the same setup as the `Shared' model, but with fewer cameras involved.
\end{enumerate}

The above-listed 5 network architectures can be broadly categorized into three groups. First, the stacked input model (3C-1N) serves as a baseline to evaluate the efficacy of different input data processing methods, as in computer vision, dual input, and stacking are common choices for handling multiple image inputs. Second, the networks with shared weights, specifically the 1C-1N and 3C-3N models, are based on the assumption that using shared weights to process the initial and final images will effectively highlight the differences. Third, the separate networks approach aims to leverage the full capabilities of CNNs to detect changes in the input images, with the individual networks operating independently. For simplicity, we use the 1C-1N model as a representative example, illustrating its detailed network architecture in Figure \ref{fig:architecture}. The primary distinction for separate networks is the non-shared weights. In the case of three cameras, this approach involves tripling the networks and concatenating a total of six visual embeddings. The network architecture for the stacked input model is relatively straightforward, with a single ResNet processing all inputs.

For the last layer of all architectures, we treat tools and actions as independent labels. Consequently, we employ a dual-head network architecture for the action-tool recognition tasks. This design allows for separate processing and classification of tool and action labels.

\subsection{Evaluation}
We applied a consistent evaluation approach across all network architectures. To do this, we trained each network from scratch for 150 epochs using a cross-entropy loss function. During training, we continually saved the model weights that achieved the highest validation set accuracy and used them for the final evaluation on the test set to prevent over-fitting. For hyper-parameter optimization, we performed a grid search covering learning rates (1e-3, 5e-4, 1e-4), batch sizes (16, 32, 64, 128), and first-block kernel sizes (3x3, 5x5, 7x7) with strides (1x1, 2x2), over the 16-way action-tool prediction task. The optimal hyper-parameters identified for each model were then used for all tasks. To reduce variance due to randomness and ensure our results were reproducible, we averaged the outcomes (i.e., accuracy) across 5 random seeds, and reported the 95\% confidence interval.

\begin{figure*}[!htb]
  \centering
  \subfigure[Tool recognition with action]{\includegraphics[ height=4.23cm]{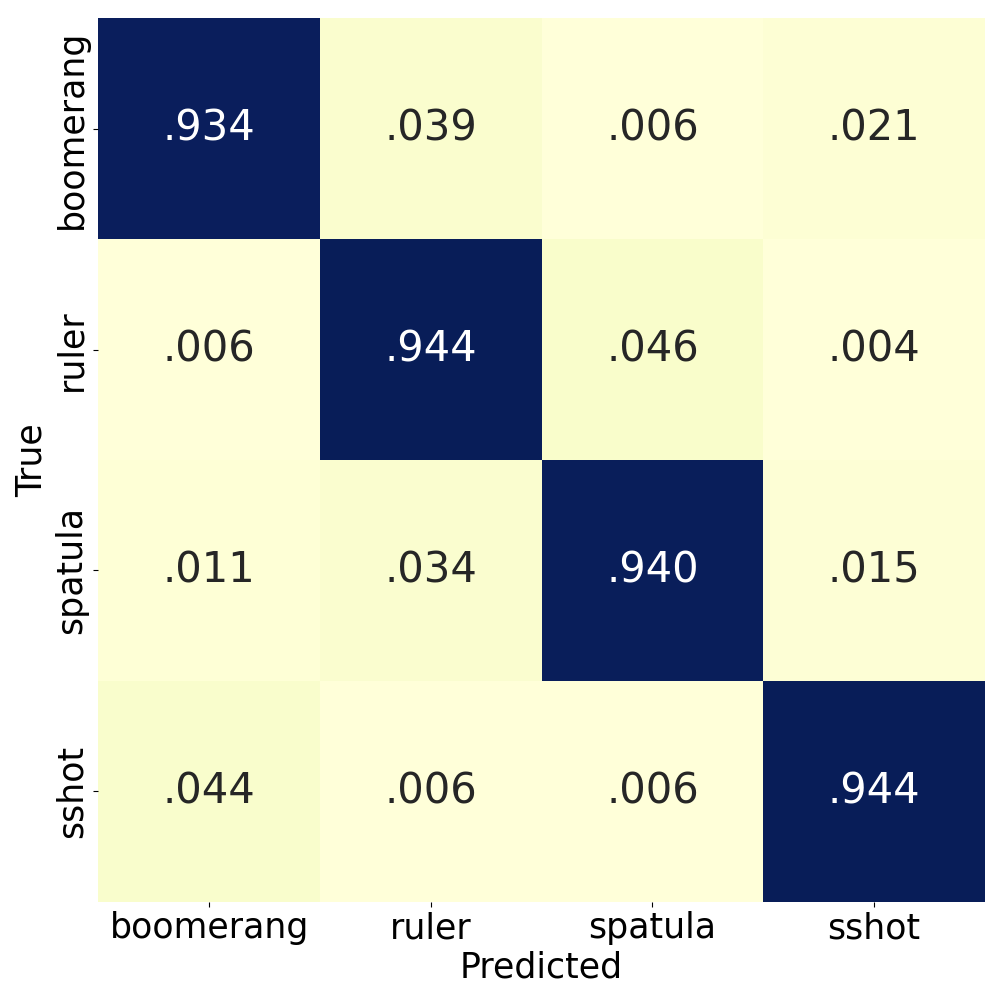}}\hspace{0.01\linewidth}
  \subfigure[Tool recognition without action]{\includegraphics[ height=4.23cm]{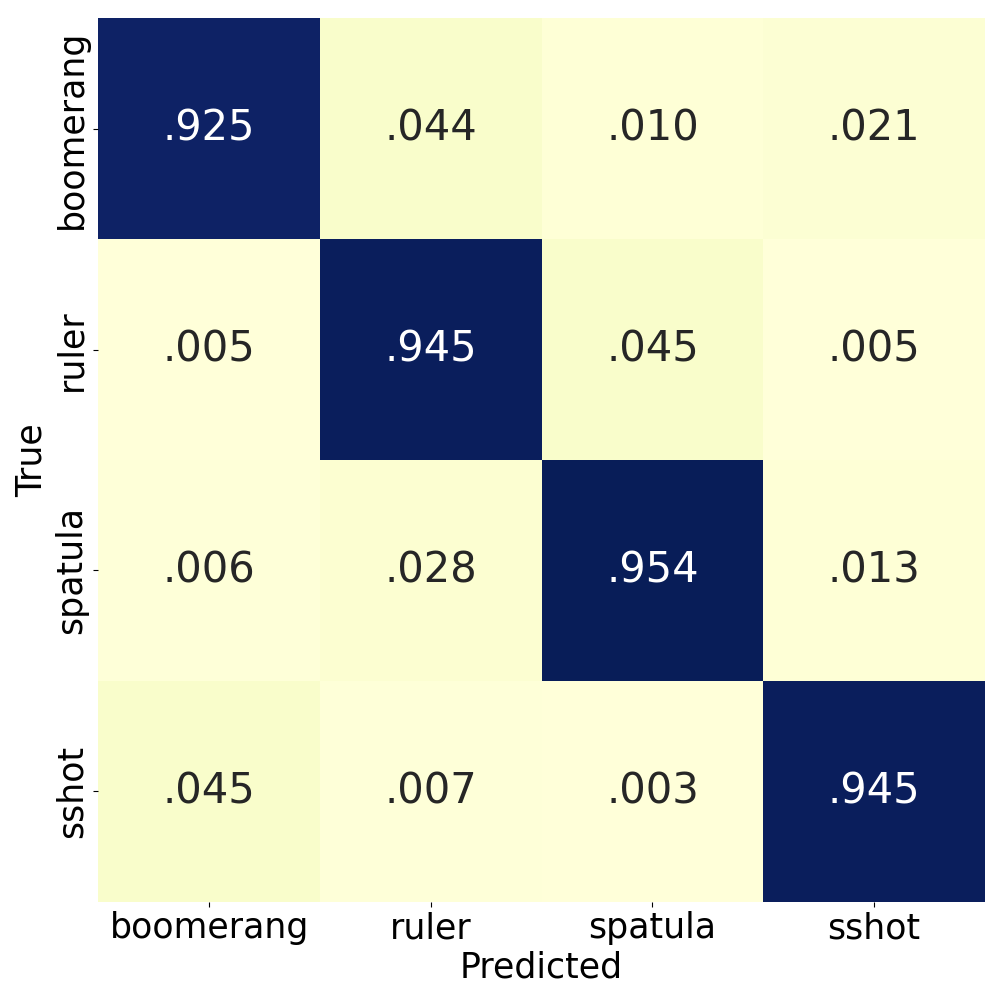}}\hspace{0.01\linewidth}
  \subfigure[`Tool' output head (tool+action prediction network) ]{\includegraphics[height=4.23cm]{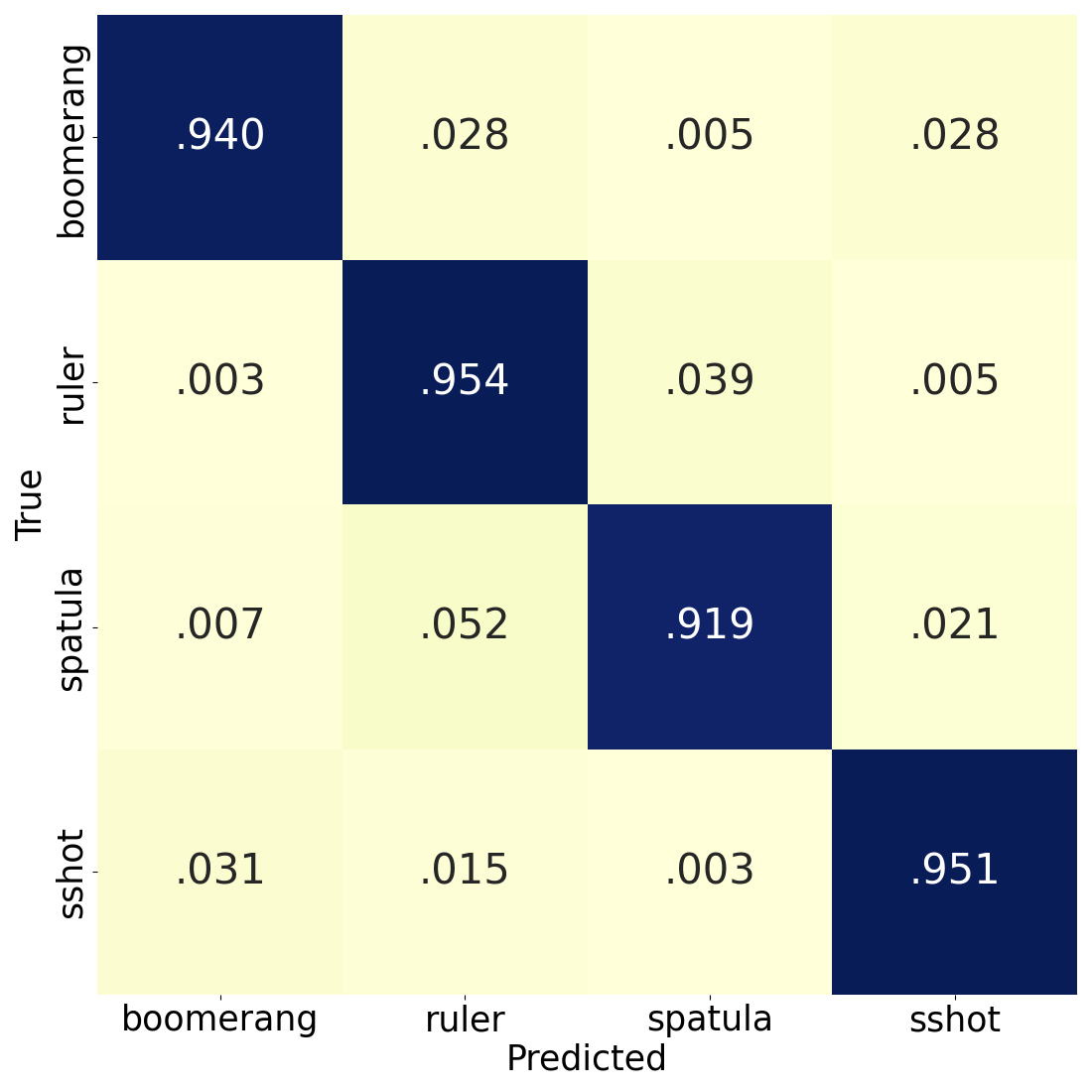}}\hspace{0.01\linewidth}
  \subfigure[`Action' output head (tool+action prediction network) ]{\includegraphics[height=4.23cm]{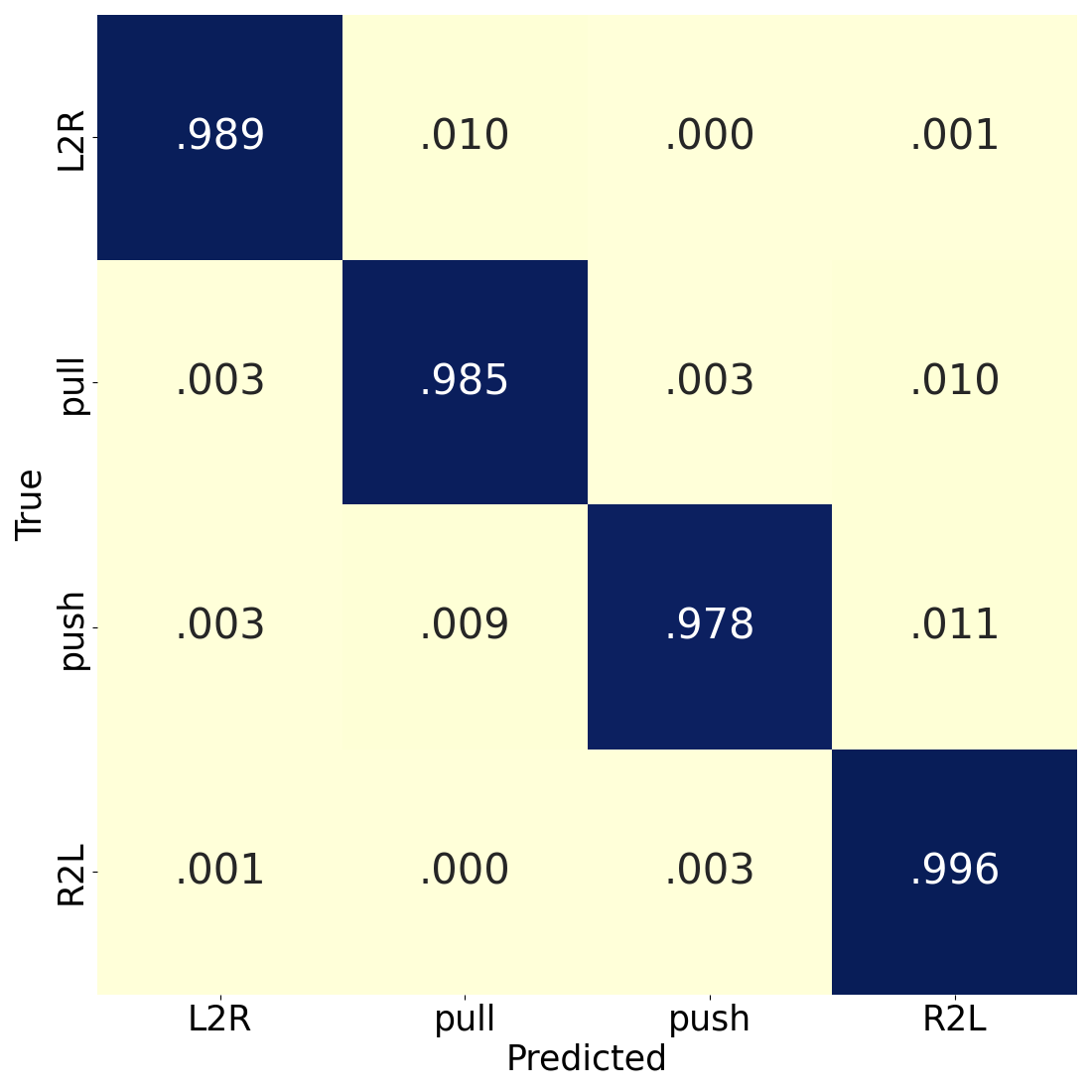}}
  \caption{Normalized Confusion Matrices for ResNet50-based 1C-1N Architecture: (a) Tool-Only Recognition with Action Reference, (b) Tool-Only Recognition without Action Reference, (c) Tool Recognition Output Head, and (d) Action Recognition Output Head. The visualization of these matrices demonstrates the model's high accuracy in different scenarios, with the action recognition component achieving near-perfect accuracy across all actions and tool recognition showing robust performance. }
  \label{fig:cf}
\end{figure*}

\section{Results} \label{results}

This section presents the results of secondary tool affordance learning. We first introduce statistics of different network architectures and the outcomes of the best-performing models. We then illustrate confusion matrices for each task to visually present the accuracy and misclassification rates, offering insights into the model's performance across various tasks.

The initial experiment aimed to evaluate the predictive accuracy of the five models on a task predicting both actions and tools. This task was selected to effectively represent secondary affordance prediction behavior using the multisensory data with three cameras from the iCub. Table \ref{netarch} presents each model's accuracy along with a 95\% confidence interval.

\begin{table}[ht!]
\centering
\caption{ Network Architectures' Accuracy and 95\% confidence interval on Action and Tool Recognition Task }
\label{netarch}
\begin{tabular}{llll}
\toprule
\textbf{Architectures} &\textbf{ResNet18}& \textbf{ResNet50} & {\textbf{ResNet101}}     \\ \midrule
Stacked-channels\\ (3C-1N)       &  76.68 ± 1.31       &   80.62 ± 2.70       &  79.00 ± 1.79                   \\ \addlinespace
Separate\\ (3C-6N)& 81.68 ± 3.35 & 73.90 ± 1.30 & 71.28 ±1.39                        \\ \addlinespace
Shared\\ (3C-3N)           & 81.56 ± 3.54 & 86.09 ± 3.02 &  80.87 ± 1.01                            \\ \addlinespace
Separate-central\\ (1C-2N)& 78.06 ± 2.57 & 76.31 ± 5.10 & 72.43 ± 1.99                            \\ \addlinespace
\textbf{Shared-central}\\ \textbf{(1C-1N)}   & \textbf{86.06 ± 2.05} & \textbf{90.78 ± 4.18} & \textbf{85.15 ± 5.24}         \\ \bottomrule
\end{tabular}
\end{table}

The results in Table \ref{netarch} above show that networks using shared-weights (3C-3N and 1C-1N) to process the initial and final images perform better than those with stacked-channel inputs (3C-1N) or separate networks (3C-6N, 1C-2N). This finding indicates that processing images independently or stacking them could interfere with comparing the initial and final states, highlighting the benefits of shared-weight networks for this task.

It is interesting to note that models that used all cameras as input (3C-* models) were found to perform worse than those that only used the central camera (1C-* models) on either separate networks or shared weights settings. We hold that this decline may be due to the inferior quality of the iCub's left and right cameras compared to the central camera.

Following the evaluation, the \textbf{1C-1N} network architecture stands out as the most effective. Therefore, we selected this network for the remaining experiments. Figure \ref{fig:architecture} depicts this architecture within the action and tool recognition task context. When predicting tools alone, action inputs are represented as one-hot vectors and combined with the visual embeddings from the ResNets, improving the precision of the secondary affordance predictions. We then applied the 1C-1N network to all three affordance recognition tasks. In the ``tools'' task, we used images of the object's initial and final poses, as well as action information. In the ``tools no actions'' scenario, we relied solely on the initial and final images for tool prediction. Lastly, in the ``tools+actions'' task, we used initial and final images to recognize combinations of tools and actions. The accuracy of each task is reported in Table \ref{1c1nresults}.

\begin{table}[ht!]
\centering
\caption{Accuracy and 95\% confidence interval of ResNet-based 1C-1N architectures in Tool Recognition Tasks}
\label{1c1nresults}
\begin{tabular}{llll}
\toprule
\textbf{Output} &\textbf{Class}& \textbf{Models} & {\textbf{Accuracy (\%)}}  \\ \midrule

Tools           &4& ResNet18        & 89.00 ± 2.21               \\
                && \textbf{ResNet50}         & \textbf{94.03 ± 1.23}   \\
                && ResNet101        & 93.06 ± 2.38               \\ \hline\addlinespace
Tools (no Actions)&4& ResNet18       & 88.78 ± 2.30              \\
                && \textbf{ResNet50 }         & \textbf{94.21 ± 0.92}    \\
                && ResNet101       & 92.50 ± 2.02              \\ \hline\addlinespace
Tools+Actions   &4+4& ResNet18      & 86.06 ± 2.05                \\
                && \textbf{ResNet50}        & \textbf{90.78 ± 4.18}    \\
                && ResNet101       & 85.15 ± 5.23              \\ \bottomrule
\end{tabular}
\end{table}

Based on the entries in Table \ref{1c1nresults}, we found that the ResNet50 architecture consistently outperformed the other two across all tasks, including the challenging task of recognizing tools and actions simultaneously.

Finally, we examine the error patterns across the three tasks by analyzing the normalized confusion matrices for the ResNet50-based 1C-1N architecture, as shown in Figure \ref{fig:cf}. The results show no obvious misclassification pattern, which may be attributable to the difference in action and tool type. However, it is evident that ``action'' prediction is easier for the network than ``tool'' prediction, as seen in panels (c) and (d).

\subsection{Supplementary results and ablation study}
To further look into the relative difficulty of recognizing actions rather than tools, we re-trained the CNNs on an action-only prediction task. The results reveal that this task can be effectively solved with more than 99\% accuracy with all models we tested (see section \ref{methods}). This highlights the following aspects of our dataset.  It is comparatively easier to discriminate between tools and actions when considered separately than to determine the joint combination of tools and actions.  This indicates that for any given action,  different tools could produce a similar result (i.e., the final pose of the object). In addition, a single tool is also suitable for multiple actions. 

We also conducted an ablation study in which we performed explicit classification over all possible combinations of the 4 actions and 4 tools, equating to 16 categories. However, this approach did not yield performance comparable to the dual-head networks, where tools and actions were classified separately. To explain this outcome, we propose that the 16-class classification may cause redundant network weights in the final layer, unlike the streamlined, dual-headed network structure.

It is important to note that in our experiments, all tools were used to perform all actions. Yet, in real-life scenarios, tools are typically not interchangeable. This means that the likelihood of actions and tools occurring together might not easily separate into distinct probabilities for each. In such cases, using a single output layer that predicts all possible action-tool combinations could be advantageous.

\section{Reproducibility of The Study} \label{reprod}
We used a public repository\footnote{https://github.com/dingdingding60/HUMANOIDS-2024} for resources --including scripts, trained models, ablation parameters, figures, and software packages with version numbers-- to reproduce the results.

\section{Conclusions} \label{conclusion}

In this work, we investigated several variants of ResNet-based deep neural network architectures and highlighted most successful ones that can serve as a subsystem for the cognitive architecture of a humanoid robot. In particular, we have shown that ResNet-based deep learning models enable the iCub robot to learn secondary tool affordances based on tool use demonstrations of human partners. These models process egocentric visual data, i.e., images capturing the initial and final positions of objects when actions are performed with a tool, to make predictions about the tools.
Based on the presented results, we suggest that interactive robots could learn to understand tool uses, aiding in collaborative tasks like furniture assembly or kitchen help.

Given that our dataset allows for the use of various tools to accomplish the same task,  we emphasize that this aspect is particularly significant for learning secondary tool affordances. Overall, our future research will continue in this direction, further exploring how to integrate multi-sensory data in a human-robot collaboration task, where the robot actively interacts with the human partners.

\bibliographystyle{IEEEtran}
\bibliography{IEEEabrv,BosongAIRLab}

\end{document}